\documentclass{article}

\usepackage{iclr2027_conference,times}

\usepackage{amsmath,amsfonts,bm}

\def\eqref#1{equation~\ref{#1}}

\def\1{\bm{1}}

\DeclareMathAlphabet{\mathsfit}{\encodingdefault}{\sfdefault}{m}{sl}
\SetMathAlphabet{\mathsfit}{bold}{\encodingdefault}{\sfdefault}{bx}{n}

\usepackage{amsmath,amssymb}
\usepackage{booktabs}
\usepackage{graphicx}
\usepackage{microtype}
\usepackage{multirow}
\usepackage{subcaption}
\usepackage{float}
\usepackage{xcolor}
\usepackage{hyperref}
\usepackage{url}

\usepackage{colortbl}
\usepackage{wrapfig}
\usepackage{mathtools}
\usepackage{arydshln}

\definecolor{todored}{RGB}{180,35,35}

\newcommand{\agent}{third-person}
\newcommand{\wrist}{wrist}
\newcommand{\liberop}{LIBERO-Plus}

\title{Does Adversarial Training Improve Generalization in Multi-View VLAs? Revealing and Mitigating View Collapse}

\author{Futa Waseda, Shuhei Kurita \& Isao Echizen \\
National Institute of Informatics\\
Tokyo, Japan \\
\texttt{\{futa-waseda, skurita, iechizen\}@nii.ac.jp} \\
}

\iclrfinalcopy

\begin{document}

\maketitle

\begin{abstract}
Vision-language-action (VLA) models adapt pretrained vision-language models (VLMs) for closed-loop robot control, transferring their perceptual and semantic capabilities to action prediction. Despite strong in-distribution performance, however, VLAs often degrade under deployment shifts. Adversarial training (AT) offers a model-adaptive approach to robustness without explicitly anticipating individual shifts, but its effect on natural distribution-shift generalization in multi-view VLAs remains unclear. We study this question using a multi-view VLA directly adapted from a pretrained VLM and evaluate generalization across seven LIBERO-Plus shift axes. Direct AT substantially improves Camera Viewpoint and Sensor Noise, the two shifts affecting only the third-person view, yet produces mixed or negative effects on other shifts. 
Controlled view interventions reveal a surprising failure mode that we term \emph{view collapse}: Direct AT can shift cross-view reliance so strongly that the policy becomes dominated by the wrist view. This exposes a \textit{robustness shortcut}: apparent robustness to a shifted view can arise from reduced use of that view rather than more robust perception of it.
This motivates a distinction between \emph{robust perception}, extracting reliable information under within-view shifts, and \emph{robust fusion}, adapting reliance across views according to their reliability. To reduce fixed view reliance, we use a simple View Swap intervention and then re-evaluate AT. With View Swap, AT further improves Camera Viewpoint, Sensor Noise, and Robot Initial State, while its effects remain mixed on other shifts. Our results show that multi-view robustness requires separating improved perception from changes in cross-view reliance, and that AT provides selective rather than generic distribution-shift benefits.
\end{abstract}

\section{Introduction}

Vision-language-action (VLA) models transfer the perceptual and semantic capabilities of pretrained vision-language models (VLMs) to closed-loop robot control and achieve high success on standardized manipulation benchmarks~\citep{kim2024openvla,kim2025oft,black2025pi05,starvla2026}.
High in-distribution performance, however, does not ensure reliable deployment.
Recent evaluations show substantial degradation under deployment shifts, such as camera viewpoints, sensor noise, lighting, and object layouts~\citep{fei2026liberoplus,zhou2025liberopro}.
A key challenge is therefore to develop VLA policies that remain reliable beyond the conditions represented in their limited robot demonstrations.

One effective approach to improving VLA robustness is to incorporate anticipated environmental variations into training through expanded datasets or predefined augmentations~\citep{fei2026liberoplus,guo2026robustness}.
However, this approach requires specifying which variations to model and how to generate them. As environments and sensors vary, anticipating and designing training examples for each potential deployment shift becomes increasingly difficult and costly.
\textit{Can we improve distribution-shift generalization while reducing the need to explicitly model individual environmental changes?}
Adversarial training (AT) offers one possible approach.
Rather than defining a fixed set of augmentations, AT generates model-adaptive perturbations that expose weaknesses of the current policy during optimization.
In computer vision, adversarially generated examples have sometimes improved robustness beyond the perturbation used for training, including generalization to unseen domains and common corruptions~\citep{volpi2018generalizing,xie2020advprop,rusak2020simple,kireev2022effectiveness}.
These benefits are not universal, however, as AT can alter feature reliance and amplify alternative shortcuts~\citep{moayeri2022tradeoffs}.
This motivates our central question:
\begin{quote}
\emph{Does adversarial training improve distribution-shift generalization in VLA models?}
\end{quote}

We investigate this question in the multi-view direct VLM-to-VLA setting, where pretrained VLMs are adapted directly to robot policies without intermediate large-scale robot-action pretraining~\citep{starvla2026,ye2026starvlaalpha,zhang2026vlm4vla,wang2026vla}.
This \emph{direct VLM-to-VLA} setting is attractive both practically and experimentally: it enables flexible adoption of new VLM backbones without repeating large-scale robot-action pretraining, while removing such pretraining as a confounding factor when studying downstream generalization.
Following common multi-view VLA designs~\citep{kim2025oft,black2025pi05,starvla2026}, our policy receives a global \agent{} view and a close-range \wrist{} view, and we evaluate generalization across seven \liberop{} shift axes without generating the corresponding evaluation conditions during training.


Surprisingly, applying AT directly from the pretrained VLM substantially improves Camera Viewpoint and Sensor Noise, both of which affect only the third-person stream, while producing mixed effects across the remaining shifts.
Controlled view interventions reveal a \textit{view-collapse failure mode} under Direct AT: cross-view reliance can shift so strongly that the policy becomes dominated by the wrist view, showing little sensitivity to degradation of the third-person view.
This exposes a \textit{robustness shortcut}: gains under shifts affecting one camera can arise from reduced use of that camera rather than more robust use of it.
This observation motivates a distinction between two complementary capabilities: (1)
\textbf{robust perception}, extracting task-relevant information despite within-view distribution shifts, and (2)
\textbf{robust fusion}, adjusting reliance across views according to their reliability.
AT may support robust perception by perturbing images, but it does not require the policy to retain useful information from both views. It therefore permits solutions that rely predominantly on one visual source when that is sufficient for minimizing the action loss.

To mitigate this failure mode, we first initialize from an SFT policy and use View Swap, a simple intervention that randomly replaces one camera view with the corresponding view from another training sample, making that view unreliable while preserving the original action target.
View Swap substantially improves Camera Viewpoint and Sensor Noise, the two shifts affecting only the third-person view, while preserving original LIBERO performance.
On top of View Swap, which reduces fixed view reliance, we re-evaluate the incremental effect of AT. Notably, the benefits of AT extend beyond image corruptions: performance under Robot Initial State improves consistently across settings, while Sensor Noise and Camera Viewpoint further improve in most cases. However, gains are smaller for Object Layout and can be negative for lighting, backgrounds, and language, showing that AT provides complementary, yet strongly shift-dependent, generalization benefits.

Our contributions are threefold:
\begin{itemize}
    \item We identify \emph{view collapse} as a failure mode of AT in VLM-to-VLA adaptation: gains under some shifts can arise from persistent single-view reliance rather than more robust perception.

    \item We distinguish \textit{robust perception} from \textit{robust fusion} as complementary aspects of multi-view robustness, and use View Swap as a simple intervention to reduce fixed view reliance.

    \item We characterize the selective transfer of AT after mitigating fixed view reliance, finding consistent gains on Camera Viewpoint, Sensor Noise, and Robot Initial State across direct VLM-to-VLA configurations, but mixed effects on other shifts.
\end{itemize}

\section{Related Work}

\textbf{VLA generalization under distribution shifts.}
Recent benchmarks expose substantial VLA performance degradation beyond standard evaluation conditions~\citep{pumacay2024colosseum,zhou2025liberopro}, with LIBERO-Plus providing a fine-grained evaluation across seven visual, linguistic, and robot-state shift axes~\citep{fei2026liberoplus}. Existing approaches improve such generalization through several routes. LIBERO-Plus uses shift-specific training data~\citep{fei2026liberoplus}, while RobustVLA~\citep{guo2026robustness} combines predefined augmentations with adversarial training. Model-centric methods improve visual processing or preserve pretrained representations through recurrent attention, backbone restoration, or architectural design~\citep{xiao2026avavla,dey2025revla,fu2025mergevla}, while other work expands policy experience through online reinforcement learning or heterogeneous co-training~\citep{fei2025srpo,ye2026starvlaalpha}. 
In contrast, we investigate AT as a model-adaptive training strategy that requires neither shift-specific training data nor modifications to the underlying VLA architecture.

\textbf{Adversarial training for generalization.}
Adversarial training (AT)~\citep{goodfellow2014explaining,madry2018towards} optimizes models against worst-case, norm-bounded input perturbations. Beyond adversarial defense, model-adaptive perturbations have been used to expand source domains, improve image recognition, and generalize to unseen corruptions~\citep{volpi2018generalizing,xie2020advprop,rusak2020simple}. Appropriately tuned $\ell_p$-bounded AT can also improve accuracy and calibration under common image corruptions~\citep{kireev2022effectiveness}. 
However, adversarial robustness can substantially alter the features learned and used by a model~\citep{tsipras2019robustness, ilyas2019adversarial}, and such changes do not necessarily improve all forms of natural robustness~\citep{moayeri2022tradeoffs}.
Recent VLA work has incorporated AT into robustness-oriented training~\citep{guo2026robustness}, while other studies focus on adversarial attacks and defenses~\citep{wang2024vlaattack,xu2025edpa}. In contrast, our focus is diagnostic: we study how AT affects natural distribution-shift generalization and reshapes multi-view reliance in VLAs.

\textbf{Multi-view policy learning.} 
Multiple camera views provide complementary information that reduces occlusion and perceptual ambiguity in manipulation~\citep{akinola2021multiview}. Wrist cameras offer stable hand-centric observations but have limited field of view, motivating their combination with global views~\citep{hsu2022hands}. 
Prior work addresses sensor failure by randomly masking views or sensor modalities during training~\citep{neverova2015moddrop,akinola2021multiview,skand2025masked}. 
Related work in multimodal learning
has also shown that models can favor one modality over others
\citep{wu2022characterizing,du2023uni}. However, how robustness-oriented
training reshapes view reliance in closed-loop multi-view policies remains
less understood.

\section{Study Setup}
\label{sec:setup}

\subsection{Multi-View VLA Training}
\label{sec:vla-training}

Let $f_\theta$ denote a multi-view VLA policy. At each timestep, it receives a
third-person observation $x^g$, a wrist-camera observation $x^w$, and a language
instruction $\ell$, and predicts an action or action chunk
\begin{equation}
    \hat{a}=f_\theta(x^g,x^w,\ell).
    \label{eq:policy}
\end{equation}
The two visual streams provide complementary observations of the same
manipulation episode: the \agent{} view captures global scene state, whereas
the \wrist{} view provides a close-range observation of the end effector and
nearby objects. 

Let $\mathcal{D}$ denote a demonstration dataset and $a$ the target action or
action chunk. Supervised fine-tuning (SFT) minimizes
\begin{equation}
    \mathcal{L}_{\mathrm{SFT}}(\theta)
    =\mathbb{E}_{(x^g,x^w,\ell,a)\sim\mathcal{D}}
    \left[
        \ell_{\mathrm{act}}
        \bigl(f_\theta(x^g,x^w,\ell),a\bigr)
    \right],
    \label{eq:sft}
\end{equation}
where $\ell_{\mathrm{act}}$ denotes the training objective associated with the
policy's action decoder.
This notation covers both direct action-regression objectives~\citep{kim2025oft} and generative objectives such as flow
matching~\citep{black2025pi0,black2025pi05}; our analysis does not depend on a particular action-decoding architecture.

\subsection{Adversarial Training}
\label{sec:adversarial-training}

Adversarial training (AT) replaces the visual inputs with model-adaptive,
norm-bounded perturbations. For both views, the inner maximization is
\begin{align}
    (\delta^{g\star},\delta^{w\star})
    \in \arg\max_{
        \|\delta^g\|_\infty\leq\epsilon,
        \|\delta^w\|_\infty\leq\epsilon}
    \ell_{\mathrm{act}}
    \bigl(
        f_\theta(x^g+\delta^g,x^w+\delta^w,\ell),a
    \bigr),
    \label{eq:inner-at}
\end{align}
yielding the training objective
\begin{equation}
    \mathcal{L}_{\mathrm{AT}}(\theta)
    =\mathbb{E}_{\mathcal{D}}
    \left[
        \ell_{\mathrm{act}}
        \bigl(
            f_\theta(x^g+\delta^{g\star},
                     x^w+\delta^{w\star},\ell), a
        \bigr)
    \right].
    \label{eq:at}
\end{equation}
Our central question is whether optimizing Eq.~\ref{eq:at} on the training
distribution improves policy success under natural distribution shifts that
are neither observed during training nor used to construct the adversarial
perturbations.

\subsection{Experimental Instantiation}
\label{sec:experimental-instantiation}

\textbf{Model and adaptation.}
We instantiate this formulation through \textit{direct VLM-to-VLA adaptation} using the public StarVLA implementation~\citep{starvla2026}, following the simple architecture studied in StarVLA-$\alpha$~\citep{ye2026starvlaalpha}.
Specifically, we directly adapt Qwen3.5-0.8B~\citep{qwen2026qwen35} to robot control without intermediate action-specific robot pretraining.
We adapt the VLM with LoRA (rank 32) and attach an OFT-style lightweight MLP head that regresses eight-step continuous action chunks~\citep{kim2025oft,ye2026starvlaalpha}.
A single policy is trained jointly across all four LIBERO suites: Spatial, Object, Goal, and Long.
Further training details are provided in Appendix~\ref{app:implementation}.


\textbf{Adversarial optimization.}
Multi-step PGD-AT~\citep{madry2018towards} substantially increases the cost of
fine-tuning large VLA policies. We therefore use Fast-AT~\citep{wong2020fast},
which approximates the inner maximization with one gradient step from a random
initialization within the perturbation set. Unless otherwise specified, we use
an $\ell_\infty$ radius of $\epsilon=1/255$, step size $\epsilon$, and a
uniform random start within the $\epsilon$-ball. Perturbations are optimized
jointly for both visual streams.

\textbf{Evaluation.}
We evaluate rollout success on original LIBERO~\citep{liu2023libero} and on the seven unseen shift axes of \liberop{}~\citep{fei2026liberoplus}: Background, Robot Initial State, Camera Viewpoint, Language, Sensor Noise, Object Layout, and Lighting.
We report each shift axis and the overall LIBERO-Plus success rate (Total).
Further details are provided in Appendix~\ref{app:implementation}.

\section{Adversarial Training Can Lead to View Collapse}
\label{sec:view-collapse}

This section examines whether pixel-space AT improves generalization to distribution shifts that are not explicitly modeled during training, and analyzes how AT affects multi-view information use.
For this analysis, both SFT and Direct AT are trained from the pretrained VLM initialization.

\subsection{Selected Shift Gains Do Not Imply Uniform Generalization}
\label{sec:at-profile}


Table~\ref{tab:direct-at-profile} summarizes the resulting performance profile. 
Direct AT substantially improves Camera Viewpoint (+48.6 pp) and Sensor Noise (+33.5 pp), while degrading Robot Initial State (-18.8 pp), Lighting (-16.6 pp), and Background (-15.5 pp). Despite these trade-offs, the overall LIBERO-Plus success improves from 62.7\% to 69.6\%, showing that aggregate gains can coexist with highly uneven shift-specific behavior.
Additional threat-model and budget ablations are provided in Table~\ref{tab:at-threat-ablation} in the Appendix, showing that similar shift-dependent behavior appears across multiple perturbation norms and budgets, although the magnitude varies with the AT configuration.


\begin{table*}[t]
\centering
\caption{\textbf{Adversarial Training (AT) preserves original-task performance and improves LIBERO-Plus performance, but its effects are highly shift-dependent.} It strongly improves Camera (camera viewpoint) and Noise (sensor noise), the two shifts affecting only the third-person view, while degrading several others. Values are four-suite success rates (\%).}
\label{tab:direct-at-profile}
\small
\resizebox{\textwidth}{!}{%
\begin{tabular}{lccccccccc}
\toprule
& \multicolumn{1}{c}{\textbf{Original}} & \multicolumn{8}{c}{\textbf{LIBERO-Plus}} \\
\cmidrule(lr){2-2}
\cmidrule(lr){3-10}
& \textbf{Total} & \textbf{Camera} & \textbf{Robot} & \textbf{Language} & \textbf{Light} & \textbf{Background} & \textbf{Noise} & \textbf{Layout} & \textbf{Total} \\
\midrule
SFT & 92.7 & 39.8 & 55.7 & 64.4 & 84.8 & 82.2 & 53.3 & 71.7 & 62.7 \\
\midrule
\textbf{Direct AT} & 92.9 & 88.4 & 36.9 & 69.2 & 68.2 & 66.7 & 86.8 & 68.6 & 69.6 \\[0.2mm]
& \cellcolor{green!5}$\uparrow 0.2$
& \cellcolor{green!40}$\uparrow 48.6$
& \cellcolor{red!20}$\downarrow 18.8$
& \cellcolor{green!8}$\uparrow 4.8$
& \cellcolor{red!18}$\downarrow 16.6$
& \cellcolor{red!17}$\downarrow 15.5$
& \cellcolor{green!30}$\uparrow 33.5$
& \cellcolor{red!6}$\downarrow 3.1$
& \cellcolor{green!10}$\uparrow 6.9$ \\
\bottomrule
\end{tabular}%
}
\end{table*}

The two largest gains share an important property: Camera Viewpoint and Sensor Noise perturb only the \agent{} stream, while leaving the \wrist{} stream unchanged. This creates a key ambiguity: \textit{did AT make perception from the \agent{} view more robust, or did the policy learn to solve these shifts by largely ignoring that view?}

\subsection{The Selected Gains Can Arise From View Collapse}
\label{sec:view-diagnosis}


\begin{figure}[t]
    \centering
    \includegraphics[width=\linewidth]{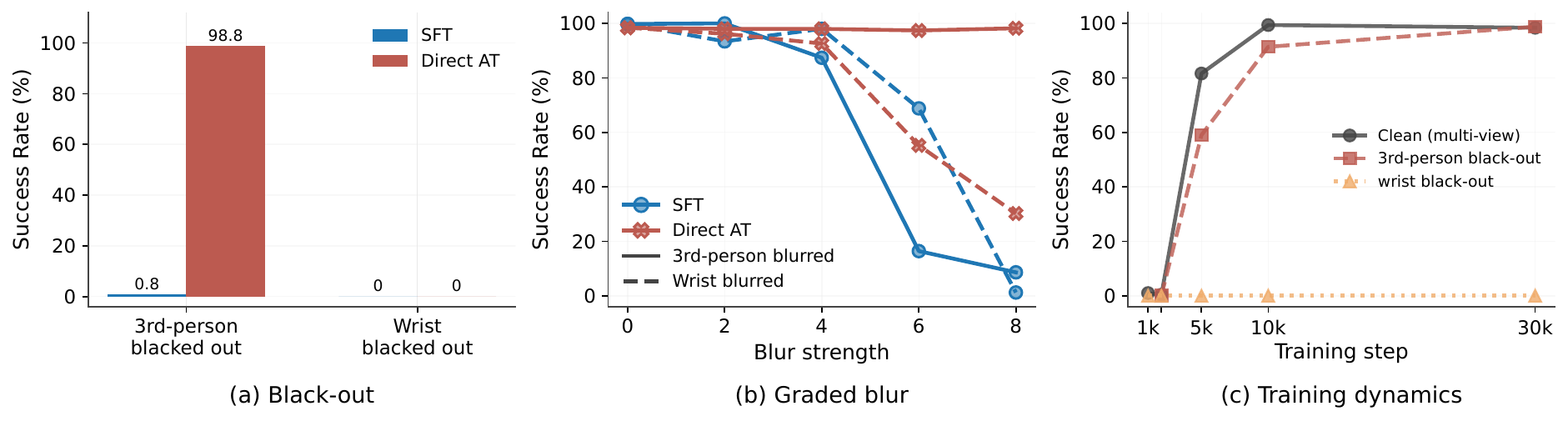}
    \vspace{-15pt}
    \caption{\textbf{Controlled interventions reveal wrist-dominant view collapse.}
    (a) Direct AT remains functional with only the \wrist{} view but fails with only the \agent{} view.
    (b) Graded blur confirms the same asymmetric reliance under progressive degradation.
    (c) Intermediate checkpoints show that wrist-dominant reliance can emerge together with task competence.
    All panels use LIBERO-Object.}
    \label{fig:viewdiag}
    \vspace{-5pt}
\end{figure}

To distinguish (i) more robust \agent{} perception from (ii) reduced use of the \agent{} view, we measure the trained policy's dependence on each visual stream. We first black out one view while leaving the other unchanged, and additionally apply graded blur to avoid relying on an extreme out-of-distribution intervention.
We evaluate interventions in Fig.~\ref{fig:viewdiag} on 500 LIBERO-Object episodes.

\noindent\textbf{Direct AT can exhibit wrist-dominant view collapse.}
The interventions reveal a striking pattern: the Direct AT policy largely ignores the \agent{} view and relies on the \wrist{} stream.
As shown in Fig.~\ref{fig:viewdiag}(a), it retains 98.8\% success when only the \wrist{} view remains, but achieves 0.0\% when only the \agent{} view remains.
In contrast, SFT fails when either view is removed, indicating reliance on both streams. Graded blur confirms the same asymmetry (Fig.~\ref{fig:viewdiag}(b)): corrupting the \agent{} stream has almost no effect, whereas corrupting the \wrist{} stream sharply degrades performance.
We term this behavior \textit{view collapse}: extreme single-view reliance in which one view dominates the policy.
This resembles previously observed modality-imbalance phenomena in multimodal learning~\citep{wu2022characterizing, du2023uni}.
The gains under Camera Viewpoint and Sensor Noise, which perturb only the \agent{} stream, can therefore arise from reduced reliance on the affected view rather than more robust use of it.
View collapse arises in three of the four AT configurations we test ($\ell_\infty$, $\epsilon{\in}\{1,2\}/255$; $\ell_2$, $\epsilon{=}2$), but not for $\ell_2$, $\epsilon{=}1$. Additional 3k-step Direct AT experiments show wrist-biased reliance across model scales, action decoders, and a robot-pretrained policy, although the severity varies (Appendix~\ref{app:cross-model-view-reliance}).


\noindent\textbf{View collapse can emerge with task competence.}
Intermediate Direct AT checkpoints show that, in this run, the policy is already wrist-dominant at the earliest checkpoint with non-trivial task success (Figure~\ref{fig:viewdiag}(c)): performance with the \agent{} view removed closely tracks clean performance, while success remains near zero when the \wrist{} view is removed. This pattern persists as task performance improves, showing that view collapse can emerge together with task competence.

\begin{wraptable}{r}{0.36\columnwidth}
    \vspace{-14pt}
    \centering
    \caption{\textbf{Matched single-view SFT.}}
    \label{tab:single-view}
    \vspace{-9pt}
    \footnotesize
    \begin{tabular}{lc}
        \toprule
        \textbf{Views} & \textbf{Success (\%)} \\
        \midrule
        Multi-view          & \textbf{92.7} \\
        \wrist{} only       & 80.6 \\
        \agent{} only       & 72.2 \\
        \bottomrule
    \end{tabular}
    \vspace{-10pt}
\end{wraptable}



Matched single-view SFT confirms that both views provide useful task information: neither single-view policy matches multi-view performance (Table~\ref{tab:single-view}). These results expose a \textit{robustness shortcut}: by relying largely on the \wrist{} stream, Direct AT can remain robust to \agent{}-only shifts without learning to robustly use that stream.

\textbf{What drives view collapse under AT?}
To first order, the $\ell_p$-robust objective with independent per-view budget $\epsilon$ adds a sensitivity penalty $\epsilon\sum_v \|\nabla_{x^v}\ell_{\mathrm{act}}\|_q$ ($q{=}1$ for $p{=}\infty$ and $q{=}2$ for $p{=}2$).
However, reducing this penalty does not necessarily require robustifying every view. If one view contributes disproportionately to the adversarial objective, AT may instead suppress reliance on that view.
We test this hypothesis with \emph{3rd-person-only AT} and \emph{wrist-only AT}, perturbing only the named view while retaining both inputs.
Interestingly, 3rd-person-only AT effectively discards the third-person view, while wrist-only AT effectively discards the wrist view.
Under the black-out evaluation, 3rd-person-only AT relies on the wrist view (88.0\% vs.\ 0.0\%), whereas wrist-only AT relies on the third-person view (50.0\% vs.\ 0.0\%), already clear at the 3k-step checkpoint.
Thus, view collapse is not specific to the third-person camera; AT may suppress reliance on a view when doing so is easier than learning to use that view robustly.
We also find asymmetric sensitivity at initialization: attacking the third-person view increases the loss more than attacking the wrist view, which may bias multi-view AT toward suppressing the third-person view (Appendix~\ref{app:view-collapse-mechanism}).

\subsection{Robust Multi-View Generalization Involves Both Perception and Fusion}
\label{sec:robustness-aspects}

The preceding experiments expose two conceptually distinct but behaviorally intertwined aspects of multi-view robustness. A policy must continue extracting useful information when observations within a view change, while also adapting how it combines multiple views as their relative reliability changes. We refer to these aspects as \emph{robust perception} and \emph{robust fusion}, respectively.

\noindent\textbf{(1) Robust perception.} 
The ability to extract reliable, task-relevant information from a view under within-view distribution shifts, while that view remains task-aligned and potentially informative.

\noindent\textbf{(2) Robust fusion.} The ability to adaptively integrate information across views as their reliability changes. When one view becomes unreliable, the policy should rely more on the remaining view without developing permanent dependence on either source.


\begin{figure}[t]
    \centering
    \includegraphics[width=0.88\linewidth]{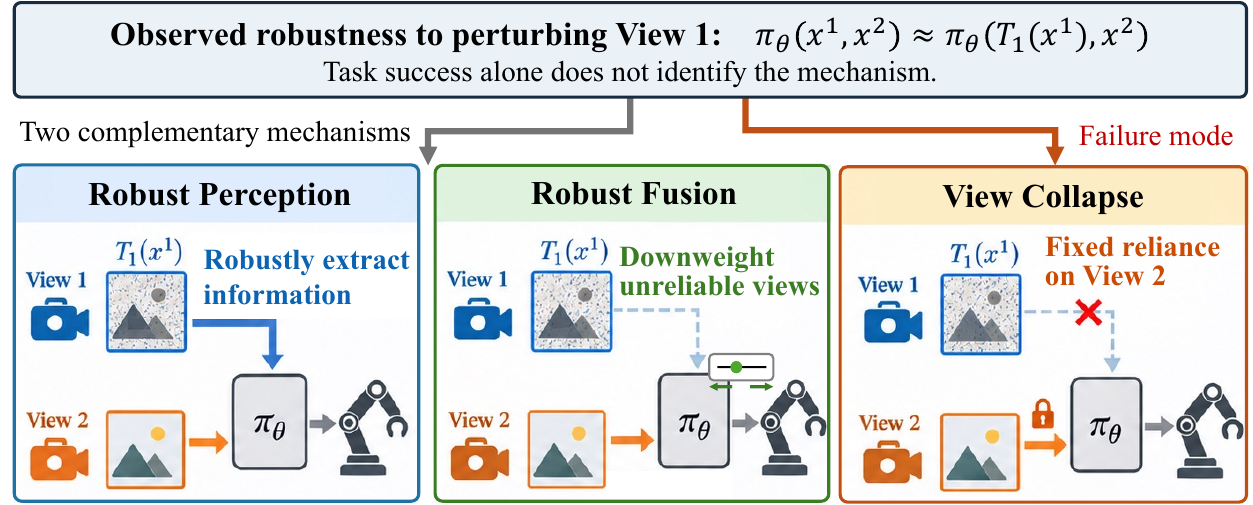}
    \caption{\textbf{Observed robustness to perturbing one view is mechanistically ambiguous.}
    Such robustness may arise from \emph{robust perception}, \emph{robust fusion}, or the failure mode of \emph{view collapse}. Here, fixed reliance means reliance that does not adapt to relative view reliability.}
    \label{fig:rob_taxonomy}
\end{figure}


View collapse illustrates that robustness to a perturbed view does not
necessarily imply robust perception of that view. When such collapse occurs, apparent robustness can result from fixed view reliance rather than improved use of the affected stream.

Pixel-space AT may contribute to robust perception by introducing perturbations within each view, but its action-level objective does not explicitly constrain cross-view utilization. It therefore permits solutions that concentrate reliance on a single stream. This motivates considering robust perception and robust fusion as two complementary aspects of multi-view robustness.

\section{Mitigating View Collapse in Adversarial Training}
\label{sec:at-with-robust-fusion}

Sec.~\ref{sec:view-collapse} shows that Direct AT can improve robustness to selected shifts through strong single-view reliance rather than robust multi-view utilization. 
How does AT behave after reducing the fixed single-view reliance? We investigate two simple strategies: SFT initialization and View Swap.

\subsection{Mitigating View Collapse: SFT Initialization and View Swap}
\label{sec:training-variants}


First, we consider \emph{SFT initialization}.
Figure~\ref{fig:viewdiag}(c) suggests that wrist-dominant reliance emerges together with task competence under Direct AT. Motivated by this observation, we first establish multi-view task behavior with standard SFT and then continue training with AT (SFT$\rightarrow$AT), aiming to reduce extreme view collapse while retaining the potential robustness benefits of AT.

Second, we consider \emph{View Swap} as a direct intervention against fixed view reliance. For a training sample $i$, we randomly select one view $s\in\{g,w\}$ and replace it with the corresponding view from another sample in the minibatch:
\begin{equation}
    \tilde{x}_i^{s} \coloneqq x_{\sigma(i)}^{s},
    \qquad
    \tilde{x}_i^{\bar{s}} \coloneqq x_i^{\bar{s}},
\end{equation}
where $\sigma$ denotes a random minibatch permutation and $\bar{s}$ denotes the unchanged view. The language instruction and action target are kept unchanged. The swapped view therefore remains visually plausible but becomes unreliable for the current task. Because either view can be swapped, the policy cannot consistently treat a particular visual source as reliable.


From the shared SFT initialization, we define four matched training variants: continued SFT, adversarial training (\textbf{SFT$\rightarrow$AT}), View Swap (\textbf{SFT$\rightarrow$Swap}), and their combination (\textbf{SFT$\rightarrow$AT+Swap}).
All variants start after 3,750 SFT steps and continue for 10k steps. View Swap is applied on 25\% of steps; SFT$\rightarrow$AT+Swap uses AT on 75\% and View Swap on 25\%.
Qualitative examples of the adversarial perturbations and View Swap intervention are provided in Appendix~\ref{app:perturbation-visualization}.

\subsection{View Interventions Reveal How Training Reshapes View Reliance}
\label{sec:view-reliance-after-mitigation}

\begin{figure}[t]
    \centering
    \includegraphics[width=0.90\linewidth]{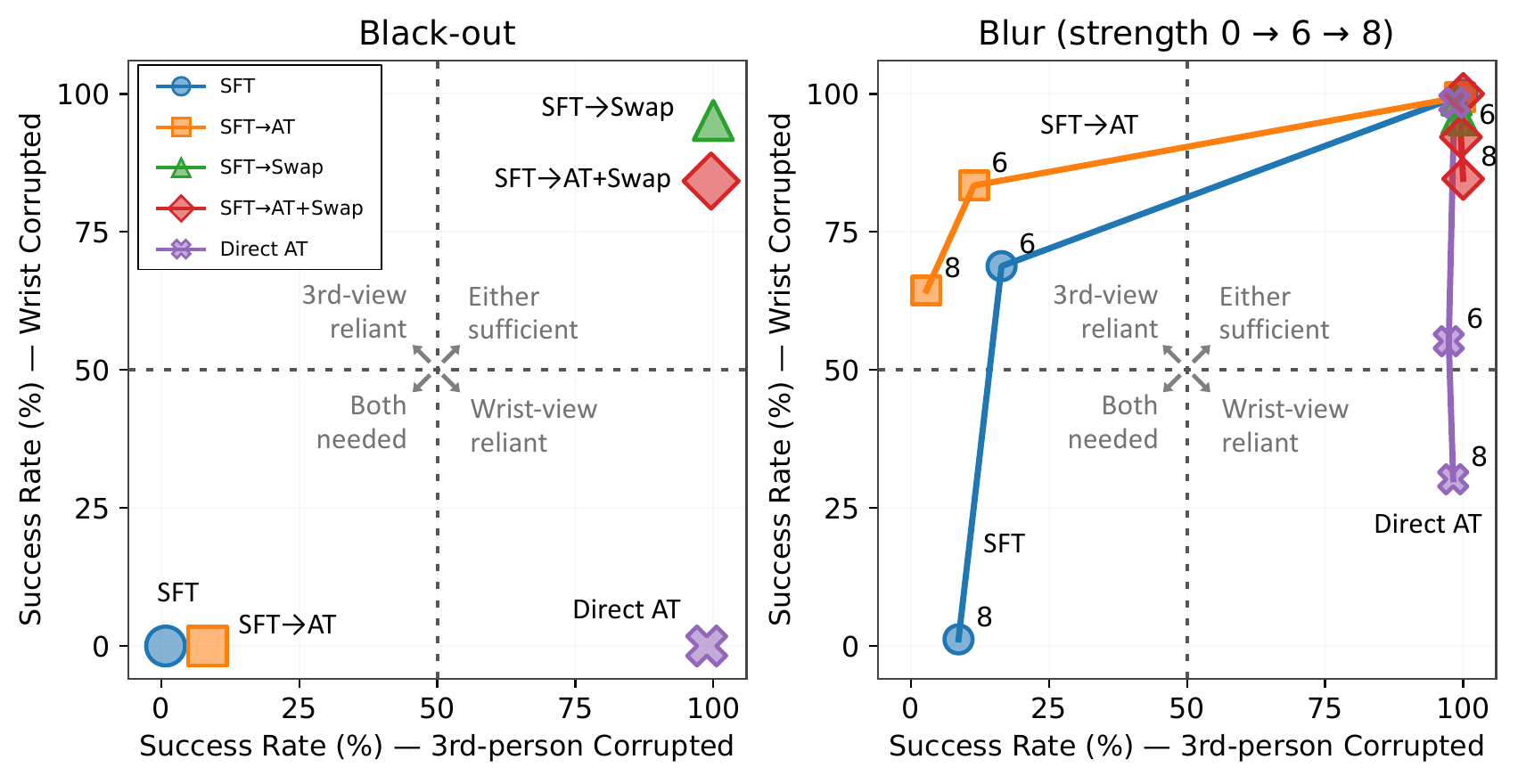}
    \vspace{-5pt}
    \caption{\textbf{Controlled view interventions reveal patterns of view reliance.}
    Left: black-out removes either the third-person or wrist view.
    Right: graded blur progressively degrades either view (strengths 0, 6, and 8).
    The horizontal and vertical axes report success under third-person- and wrist-view corruption, respectively, with the upper-right indicating flexible fallback on either view. 
    }
    \label{fig:view-reliance-map}
\end{figure}


We first examine how the training variants reshape reliance on each visual source using the same controlled view interventions as in Sec.~\ref{sec:view-collapse}. Figure~\ref{fig:view-reliance-map} summarizes the results as a two-dimensional view-reliance map, with Direct AT included for comparison.

\noindent\textbf{SFT-initialized AT does not exhibit the extreme collapse of Direct AT.}
Direct AT retains 98.8\% success when the third-person view is blacked out but fails completely when the wrist view is removed (0.0\%). In contrast, SFT$\rightarrow$AT achieves 8.4\% and 0.0\%, respectively.
Thus, the SFT-initialized policy does not exhibit the extreme wrist-dominant collapse observed under Direct AT, although its view reliance remains asymmetric.

\noindent\textbf{When introduced after SFT, AT can amplify existing view-reliance asymmetry.}
Under graded blur, SFT is more sensitive to \agent{}-view degradation than to \wrist{}-view degradation, and SFT$\rightarrow$AT further accentuates this asymmetry. This suggests that AT does not induce a universal preference for a particular camera, but can reinforce view reliance already present in the policy.

\noindent\textbf{View Swap enables flexible fallback across views.}
The behavior changes substantially when View Swap is introduced. Under black-out, SFT$\rightarrow$Swap and SFT$\rightarrow$AT+Swap remain successful when either view is removed, moving from the ``both needed'' or single-view-reliant regimes toward the ``either sufficient'' regime. The same pattern persists under graded blur: both variants remain robust as either visual source is progressively degraded. 
These results indicate that View Swap mitigates fixed view reliance and enables flexible fallback across views.


\begin{table*}[t]
\centering
\caption{\textbf{Comparison of robust adaptation strategies for Qwen3.5-0.8B-OFT under matched training steps.} Four-suite success rates (\%); green/red mark gains/losses over the SFT baseline.}
\label{tab:reeval-main}
\small
\resizebox{\textwidth}{!}{%
\begin{tabular}{lccccccccc}
\toprule
& \multicolumn{1}{c}{\textbf{Original}} & \multicolumn{8}{c}{\textbf{LIBERO-Plus}} \\
\cmidrule(lr){2-2}
\cmidrule(lr){3-10}
\textbf{Training} & \textbf{Total} & \textbf{Camera} & \textbf{Robot} & \textbf{Language} & \textbf{Light} & \textbf{Background} & \textbf{Noise} & \textbf{Layout} & \textbf{Total} \\
\midrule
SFT & 95.6 & 38.8 & 55.7 & 71.0 & \textbf{88.9} & 88.5 & 59.1 & 69.8 & 65.3 \\
\hdashline
\addlinespace[1pt]
SFT$\rightarrow$AT & \textbf{97.8} & 35.8 & 55.9 & \textbf{71.8} & 86.2 & 86.8 & 56.8 & 75.7 & 65.0 \\
& \cellcolor{green!6}$\uparrow 2.2$ & \cellcolor{red!7}$\downarrow 3.0$ & \cellcolor{green!3}$\uparrow 0.2$ & \cellcolor{green!3}$\uparrow 0.8$ & \cellcolor{red!6}$\downarrow 2.7$ & \cellcolor{red!4}$\downarrow 1.7$ & \cellcolor{red!5}$\downarrow 2.3$ & \cellcolor{green!12}$\uparrow 5.8$ & \cellcolor{red!3}$\downarrow 0.3$ \\
\hdashline
\addlinespace[1pt]
SFT$\rightarrow$Swap & 96.3 & 77.9 & 57.9 & 70.5 & 87.5 & \textbf{88.8} & 85.7 & 77.0 & 77.0 \\
& \cellcolor{green!3}$\uparrow 0.7$ & \cellcolor{green!38}$\uparrow 39.0$ & \cellcolor{green!6}$\uparrow 2.3$ & \cellcolor{red!3}$\downarrow 0.5$ & \cellcolor{red!4}$\downarrow 1.4$ & \cellcolor{green!3}$\uparrow 0.3$ & \cellcolor{green!28}$\uparrow 26.6$ & \cellcolor{green!15}$\uparrow 7.2$ & \cellcolor{green!20}$\uparrow 11.7$ \\
\hdashline
\addlinespace[1pt]
SFT$\rightarrow$AT+Swap & 96.5 & \textbf{84.1} & \textbf{65.2} & 68.4 & 88.8 & 87.7 & \textbf{88.3} & \textbf{79.5} & \textbf{79.7} \\
& \cellcolor{green!3}$\uparrow 0.9$ & \cellcolor{green!45}$\uparrow 45.2$ & \cellcolor{green!18}$\uparrow 9.5$ & \cellcolor{red!6}$\downarrow 2.5$ & \cellcolor{red!2}$\downarrow 0.1$ & \cellcolor{red!3}$\downarrow 0.7$ & \cellcolor{green!31}$\uparrow 29.2$ & \cellcolor{green!18}$\uparrow 9.6$ & \cellcolor{green!24}$\uparrow 14.3$ \\
\bottomrule
\end{tabular}%
}
\end{table*}

\subsection{Distribution-Shift Effects of View Swap and Adversarial Training}
\label{sec:shift-generalization}

We next evaluate how the observed changes in view reliance affect natural distribution-shift generalization on \liberop{}. Table~\ref{tab:reeval-main} compares the four variants across seven shift categories and the original LIBERO tasks.

\noindent\textbf{SFT initialization alone is insufficient for AT to improve overall generalization.}
The SFT$\rightarrow$AT policy does not exhibit the extreme view collapse observed under Direct AT, but does not improve aggregate \liberop{} performance. Its effects remain shift-dependent, with a gain on Layout (+5.8 pp) but degradations on Camera ($-3.0$ pp), Light ($-2.7$ pp), and Noise ($-2.3$ pp). Thus, the absence of extreme collapse alone is insufficient for AT to yield broader distribution-shift gains.

\noindent\textbf{View Swap improves robustness to third-person-view perturbations.}
SFT$\rightarrow$Swap substantially improves Camera (+39.0 pp) and Noise (+26.6 pp), the two shifts affecting only the third-person view, while preserving original LIBERO performance. Together with the view-intervention results in Sec.~\ref{sec:view-reliance-after-mitigation}, these gains are consistent with more flexible reliance across views.
With AT, a matched View Dropout baseline is less effective (Appendix~\ref{sec:swap-dropout-ablation}); one possible factor is that View Swap preserves realistic per-view image statistics, whereas View Dropout introduces an extreme missing-view pattern.


\noindent\textbf{AT provides complementary but shift-dependent gains with View Swap.}
Relative to SFT$\rightarrow$Swap, SFT$\rightarrow$AT+Swap improves Camera (+6.2 pp), Robot Initial State (+7.3), Noise (+2.6), and Layout (+2.5), preserving original LIBERO performance but yielding small or negative effects elsewhere.
Camera, Robot Initial State, and Noise improve across all four direct VLM-to-VLA configurations (Tables~\ref{tab:reeval-main} and~\ref{tab:architecture-main}).
Noise gains and Lighting degradations may partly reflect AT's frequency-dependent trade-offs: improved robustness to several blur corruptions but reduced robustness to fog and contrast~\citep{yin2019fourier}.
Camera and Robot Initial State gains are less directly explained by pixel-level robustness, which need not transfer to spatial transformations~\citep{engstrom2019exploring}.
One possibility is that AT learns more robust task-relevant features, such as hand--object spatial cues, that support control under these shifts.

\noindent\textbf{A lightweight direct VLM-to-VLA model achieves competitive overall robustness.}
Despite these shift-dependent effects, SFT$\rightarrow$AT+Swap reaches 79.7\% overall success on LIBERO-Plus, competitive with existing LIBERO-finetuned VLAs. Notably, this performance comes from directly adapting a 0.8B VLM without large-scale robot pretraining; full comparisons appear in Appendix~\ref{app:sota-libero-plus}.

\subsection{Generality of View Swap and Selective AT Transfer Across Models}
\label{sec:generality}

\begin{table*}[t]
\centering
\caption{\textbf{Comparison of adaptation strategies across model configurations.} Values are four-suite success rates (\%); green/red mark gains/losses over matched SFT baselines.}
\label{tab:architecture-main}
\small
{\renewcommand{\arraystretch}{0.96}
\resizebox{0.92\textwidth}{!}{%
\begin{tabular}{lccccccccc}
\toprule
& \multicolumn{1}{c}{\textbf{Original}} & \multicolumn{8}{c}{\textbf{LIBERO-Plus}} \\
\cmidrule(lr){2-2}
\cmidrule(lr){3-10}
\textbf{Training} & \textbf{Total} & \textbf{Camera} & \textbf{Robot} & \textbf{Language} & \textbf{Light} & \textbf{Background} & \textbf{Noise} & \textbf{Layout} & \textbf{Total} \\

\midrule
\multicolumn{10}{l}{\textit{Direct VLM-to-VLA adaptation}} \\
\multicolumn{10}{l}{\textbf{Qwen3.5-0.8B-PI}} \\
SFT & 95.5 & 27.7 & 51.5 & \textbf{70.5} & 80.5 & 61.2 & 54.3 & 71.3 & 58.4 \\
\hdashline
\addlinespace[1pt]
SFT$\rightarrow$Swap & 97.2 & 43.2 & 56.9 & 55.4 & \textbf{85.6} & 47.7 & 77.5 & 73.4 & 62.5 \\
& \cellcolor{green!4}$\uparrow 1.7$ & \cellcolor{green!18}$\uparrow 15.4$ & \cellcolor{green!8}$\uparrow 5.4$ & \cellcolor{red!15}$\downarrow 15.1$ & \cellcolor{green!8}$\uparrow 5.1$ & \cellcolor{red!14}$\downarrow 13.6$ & \cellcolor{green!25}$\uparrow 23.1$ & \cellcolor{green!5}$\uparrow 2.2$ & \cellcolor{green!8}$\uparrow 4.1$ \\
\hdashline
\addlinespace[1pt]
SFT$\rightarrow$AT+Swap & \textbf{98.1} & \textbf{68.8} & \textbf{64.8} & 66.7 & 71.8 & \textbf{65.9} & \textbf{82.7} & \textbf{78.0} & \textbf{71.5} \\
& \cellcolor{green!6}$\uparrow 2.6$ & \cellcolor{green!42}$\uparrow 41.1$ & \cellcolor{green!16}$\uparrow 13.3$ & \cellcolor{red!7}$\downarrow 3.8$ & \cellcolor{red!10}$\downarrow 8.7$ & \cellcolor{green!7}$\uparrow 4.6$ & \cellcolor{green!30}$\uparrow 28.4$ & \cellcolor{green!11}$\uparrow 6.7$ & \cellcolor{green!22}$\uparrow 13.1$ \\
\midrule

\multicolumn{10}{l}{\textbf{Qwen3.5-2B-OFT}} \\
SFT & \textbf{96.6} & 42.0 & 54.3 & \textbf{84.0} & \textbf{90.9} & \textbf{91.0} & 70.3 & 81.0 & 71.6 \\
\hdashline
\addlinespace[1pt]
SFT$\rightarrow$Swap & 95.8 & 81.2 & 55.1 & 73.9 & 86.2 & 84.5 & 85.3 & \textbf{81.6} & 77.7 \\
& \cellcolor{red!3}$\downarrow 0.8$ & \cellcolor{green!40}$\uparrow 39.2$ & \cellcolor{green!3}$\uparrow 0.8$ & \cellcolor{red!12}$\downarrow 10.1$ & \cellcolor{red!8}$\downarrow 4.7$ & \cellcolor{red!9}$\downarrow 6.5$ & \cellcolor{green!18}$\uparrow 15.0$ & \cellcolor{green!3}$\uparrow 0.6$ & \cellcolor{green!11}$\uparrow 6.1$ \\
\hdashline
\addlinespace[1pt]
SFT$\rightarrow$AT+Swap & 95.8 & \textbf{85.0} & \textbf{57.6} & 68.8 & 83.9 & 85.2 & \textbf{90.1} & 79.2 & \textbf{78.1} \\
& \cellcolor{red!3}$\downarrow 0.8$ & \cellcolor{green!44}$\uparrow 43.0$ & \cellcolor{green!6}$\uparrow 3.3$ & \cellcolor{red!16}$\downarrow 15.2$ & \cellcolor{red!10}$\downarrow 7.0$ & \cellcolor{red!9}$\downarrow 5.8$ & \cellcolor{green!22}$\uparrow 19.8$ & \cellcolor{red!5}$\downarrow 1.8$ & \cellcolor{green!12}$\uparrow 6.5$ \\
\midrule

\multicolumn{10}{l}{\textbf{PaliGemma-OFT}} \\
SFT & 80.9 & 12.3 & 28.7 & \textbf{36.1} & \textbf{61.6} & 30.8 & 17.4 & 48.4 & 32.4 \\
\hdashline
\addlinespace[1pt]
SFT$\rightarrow$Swap & 84.2 & 31.3 & 33.4 & 34.3 & 57.7 & 43.3 & 30.7 & 55.2 & 39.9 \\
& \cellcolor{green!6}$\uparrow 3.3$ & \cellcolor{green!21}$\uparrow 19.0$ & \cellcolor{green!8}$\uparrow 4.7$ & \cellcolor{red!4}$\downarrow 1.8$ & \cellcolor{red!7}$\downarrow 3.9$ & \cellcolor{green!15}$\uparrow 12.5$ & \cellcolor{green!16}$\uparrow 13.3$ & \cellcolor{green!11}$\uparrow 6.8$ & \cellcolor{green!11}$\uparrow 7.5$ \\
\hdashline
\addlinespace[1pt]
SFT$\rightarrow$AT+Swap & \textbf{93.6} & \textbf{74.7} & \textbf{39.1} & 36.0 & 27.8 & \textbf{56.8} & \textbf{68.6} & \textbf{65.2} & \textbf{53.6} \\
& \cellcolor{green!15}$\uparrow 12.7$ & \cellcolor{green!50}$\uparrow 62.4$ & \cellcolor{green!13}$\uparrow 10.4$ & \cellcolor{red!3}$\downarrow 0.1$ & \cellcolor{red!35}$\downarrow 33.8$ & \cellcolor{green!28}$\uparrow 26.0$ & \cellcolor{green!49}$\uparrow 51.2$ & \cellcolor{green!20}$\uparrow 16.8$ & \cellcolor{green!24}$\uparrow 21.2$ \\

\midrule
\multicolumn{10}{l}{\textit{Robot-pretrained VLA}} \\
\multicolumn{10}{l}{\textbf{$\pi_0$}} \\
SFT & \textbf{93.9} & 9.1 & 5.0 & \textbf{62.3} & \textbf{88.4} & \textbf{85.2} & 12.9 & \textbf{73.8} & \textbf{44.3} \\
\hdashline
\addlinespace[1pt]
SFT$\rightarrow$Swap & 92.3 & \textbf{16.3} & 5.1 & 51.8 & 83.5 & 77.0 & \textbf{14.4} & 66.4 & 41.5 \\
& \cellcolor{red!4}$\downarrow 1.6$
& \cellcolor{green!11}$\uparrow 7.2$
& \cellcolor{green!3}$\uparrow 0.1$
& \cellcolor{red!12}$\downarrow 10.5$
& \cellcolor{red!8}$\downarrow 4.9$
& \cellcolor{red!11}$\downarrow 8.2$
& \cellcolor{green!4}$\uparrow 1.5$
& \cellcolor{red!10}$\downarrow 7.4$
& \cellcolor{red!6}$\downarrow 2.8$ \\
\hdashline
\addlinespace[1pt]
SFT$\rightarrow$AT+Swap & 93.2 & 13.9 & \textbf{7.6} & 50.1 & 85.3 & 79.7 & 14.0 & 68.3 & 41.9 \\
& \cellcolor{red!3}$\downarrow 0.7$
& \cellcolor{green!8}$\uparrow 4.8$
& \cellcolor{green!5}$\uparrow 2.6$
& \cellcolor{red!14}$\downarrow 12.2$
& \cellcolor{red!6}$\downarrow 3.1$
& \cellcolor{red!9}$\downarrow 5.5$
& \cellcolor{green!4}$\uparrow 1.1$
& \cellcolor{red!9}$\downarrow 5.5$
& \cellcolor{red!5}$\downarrow 2.4$ \\

\bottomrule
\end{tabular}%
}}
\end{table*}

We next test whether View Swap and the selective transfer of AT generalize across action decoders, model scales, VLM backbones, and pretraining regimes. Beyond the primary Qwen3.5-0.8B-OFT setting, we evaluate Qwen3.5-0.8B with a PI-style decoder, Qwen3.5-2B and PaliGemma~\citep{beyer2024paligemma} with the same OFT-style decoder, and the robot-pretrained $\pi_0$ policy (Table~\ref{tab:architecture-main}).

\noindent\textbf{View Swap consistently improves Camera Viewpoint and Sensor Noise across the direct VLM-to-VLA configurations.}
These improvements hold across Qwen model scales and action decoders as well as the PaliGemma backbone, consistent with the link observed in the primary setting between reducing fixed view reliance and robustness to view-specific shifts.

\noindent\textbf{AT shows consistent but selective transfer across direct VLM-to-VLA configurations.} 
On top of View Swap, AT further improves Camera Viewpoint, Sensor Noise, and Robot Initial State in all four configurations, while effects on the remaining shifts are mixed. 
PaliGemma shows particularly large Camera and Noise gains but a substantial Lighting degradation. 

\noindent\textbf{Pretrained $\pi_0$ exhibits a different pattern.}
At baseline, $\pi_0$ is already substantially weaker on Camera Viewpoint and Sensor Noise. View Swap partially improves these shifts, but the gains are modest, and adding AT provides no further improvement. This suggests that substantially lower starting robustness may limit recovery from View Swap and the additional benefit of AT.


\section{Discussion and Limitations}
\vspace{-3pt}


\noindent\textbf{Training trajectory and view reliance.}
Our results suggest that \textit{when} AT is introduced can affect view reliance, with extreme collapse arising under some configurations. Fully explaining the optimization dynamics underlying these view-reliance patterns remains future work.

\noindent\textbf{Robust fusion and source utilization.}
View Swap mitigates persistent single-view reliance. However, robust fusion more broadly requires adapting reliance to the relative reliability of both views, which remains to be characterized more directly.

\noindent\textbf{Scope of evaluation.}
Our controlled view-reliance analysis focuses on direct VLM-to-VLA adaptation in LIBERO/LIBERO-Plus. The different behavior of robot-pretrained $\pi_0$ suggests that pretraining may affect downstream AT benefits; broader VLAs and real-world settings remain future work.

\section{Conclusion}
\label{sec:conclusion}
\vspace{-3pt}
We studied AT for natural distribution-shift generalization in multi-view VLAs. Direct AT can reshape cross-view reliance and, in extreme cases, produce view collapse, so gains on view-specific shifts need not reflect robust use of the perturbed view. This motivates distinguishing robust perception from robust fusion. With View Swap reducing fixed view reliance, AT provides additional but strongly shift-dependent gains. Multi-view robustness should therefore consider not only task success, but also how policies use and combine views.

\section*{Reproducibility Statement}
We provide the experimental setup and major training configurations in the main paper and Appendix, including model adaptation details, adversarial-training configurations, training schedules, evaluation protocols, and hardware/GPU settings. Additional implementation details and ablations are also reported in the Appendix. We will publicly release the training and evaluation code, together with the corresponding configurations, to facilitate reproduction of our results.

\section*{Ethics Statement}
This work studies robustness and failure modes of Vision-Language-Action policies in simulated robotic environments and does not involve human participants, personal data, or sensitive information. Although we use adversarial perturbations, our goal is to diagnose robustness failures and improve the reliability of embodied AI systems rather than to enable harmful attacks. Our results are obtained in simulation and should not be interpreted as guaranteeing safety in real-world robotic deployment.

\section*{AI Use Statement}
Generative AI tools were used to assist with language editing, code development and debugging, and figure-layout ideation. All generated content was reviewed and revised by the authors. AI tools were not used to generate experimental results or determine the scientific conclusions of the paper, and all reported results, claims, and citations were verified by the authors.

\bibliography{iclr2027_conference}
\bibliographystyle{iclr2027_conference}

\clearpage

\appendix

\section{Implementation Details}
\label{app:implementation}

\subsection{Training Hyperparameters}

Tables~\ref{tab:model-optimization-config} and~\ref{tab:training-config}
summarize the implementation details for our primary Qwen3.5-0.8B-OFT
experiments.

\begin{table}[H]
    \centering
    \caption{\textbf{Model and optimization configuration for the primary setting.}}
    \label{tab:model-optimization-config}
    \small
    \setlength{\tabcolsep}{4pt}
    \begin{tabular}{p{0.36\linewidth}p{0.56\linewidth}}
        \toprule
        \textbf{Component} & \textbf{Configuration} \\
        \midrule
        VLM backbone & Qwen3.5-0.8B \\
        Action head & OFT-style lightweight MLP head \\
        Action chunk & 8 steps \\
        Input resolution & $224\times224$ raw frames; internally resized to $256\times256$ \\
        VLM adaptation & LoRA \\
        LoRA rank / $\alpha$ / dropout & 32 / 64 / 0.05 \\
        Optimizer & AdamW \\
        VLM LoRA learning rate & $1\times10^{-4}$ \\
        Action-head learning rate & $2\times10^{-5}$ \\
        Effective batch size & 128 \\
        LR schedule & Cosine decay \\
        Warmup & 300 steps \\
        Precision & bfloat16 mixed precision \\
        \bottomrule
    \end{tabular}
\end{table}

LoRA is applied within \texttt{qwen\_vl\_interface} to
\texttt{q\_proj}, \texttt{k\_proj}, \texttt{v\_proj}, \texttt{o\_proj},
\texttt{gate\_proj}, \texttt{up\_proj}, \texttt{down\_proj}, \texttt{qkv},
\texttt{proj}, \texttt{linear\_fc1}, and \texttt{linear\_fc2}.
The action head is optimized separately.

\begin{table}[H]
\centering
\caption{\textbf{Training configurations for Qwen3.5-0.8B-OFT.} Sec.~\ref{sec:view-collapse} compares SFT and Direct AT from the pretrained VLM initialization, whereas Sec.~\ref{sec:at-with-robust-fusion} compares matched continuations from a shared SFT checkpoint.}
\label{tab:training-config}
\resizebox{\linewidth}{!}{
\begin{tabular}{lcccccc}
\toprule
\textbf{Experiment} & \textbf{Initialization} & \textbf{Objective} & \textbf{Steps} & \textbf{AT norm} & \textbf{$\epsilon$} & \textbf{AT / Swap} \\
\midrule
Sec.~\ref{sec:view-collapse}: SFT & Pretrained VLM & SFT & 30k & -- & -- & -- \\
Sec.~\ref{sec:view-collapse}: Direct AT & Pretrained VLM & Fast-AT & 30k & $\ell_\infty$ & $1/255$ & 100\% / 0\% \\
\midrule
Sec.~\ref{sec:at-with-robust-fusion}: SFT & SFT (3,750 steps) & SFT & +10k & -- & -- & -- \\
Sec.~\ref{sec:at-with-robust-fusion}: SFT$\rightarrow$AT & SFT (3,750 steps) & Fast-AT & +10k & $\ell_\infty$ & $1/255$ & 100\% / 0\% \\
Sec.~\ref{sec:at-with-robust-fusion}: SFT$\rightarrow$Swap & SFT (3,750 steps) & View Swap & +10k & -- & -- & 0\% / 25\% \\
Sec.~\ref{sec:at-with-robust-fusion}: SFT$\rightarrow$AT+Swap & SFT (3,750 steps) & Fast-AT + View Swap & +10k & $\ell_\infty$ & $1/255$ & 75\% / 25\% \\
\bottomrule
\end{tabular}
}
\end{table}

For all Fast-AT experiments, we use one gradient step from a uniform random
initialization within the perturbation set, with step size equal to
$\epsilon$. Unless otherwise specified, perturbations are optimized jointly over both visual views, and
the perturbation budget is linearly ramped from zero to its target value over
the first 500 training steps.

The effective batch size of 128 is kept fixed across the two-GPU
configuration (32 samples per GPU with two-step gradient accumulation) and
the four-GPU configuration (32 samples per GPU without gradient
accumulation).

\paragraph{Model-specific training schedules.}
For the matched comparisons in Sec.~\ref{sec:at-with-robust-fusion}, the number of initial SFT optimizer updates varies across model configurations (Table~\ref{tab:additional-model-initialization}).
Within each configuration, all compared variants share the same initial checkpoint and undergo 10k further optimizer updates with the same learning-rate schedule.
Thus, training budgets are matched within each configuration, while initial SFT budgets differ across configurations.

\begin{table}[H]
\centering
\caption{\textbf{Training schedules for the matched adaptation comparisons.}
All counts denote optimizer updates. Initial SFT refers to LIBERO training before the compared variants begin.}
\label{tab:additional-model-initialization}
\small
\begin{tabular}{lrr}
\toprule
\textbf{Model} & \textbf{Initial SFT updates} & \textbf{Further updates} \\
\midrule
Qwen3.5-0.8B-OFT & 3,750 & 10,000 \\
Qwen3.5-0.8B-PI  & 7,500 & 10,000 \\
Qwen3.5-2B-OFT   & 15,000 & 10,000 \\
PaliGemma-OFT    & 20,000 & 10,000 \\
$\pi_0$         & 30,000 & 10,000 \\
\bottomrule
\end{tabular}
\end{table}

For $\pi_0$, we use the StarVLA-converted OpenPI LIBERO checkpoint; its official fine-tuning configuration specifies 30k updates.
Differences in improvement magnitudes across configurations should therefore not be attributed solely to architecture or model scale, since their training histories also differ.

\subsection{Evaluation Protocol}
\label{app:evaluation}
We evaluate a single policy jointly trained across the four LIBERO suites:
Spatial, Object, Goal, and Long.
Original LIBERO performance and LIBERO-Plus performance are reported
separately.
For LIBERO-Plus, we evaluate the seven shift categories used throughout the
main paper: Camera Viewpoint, Robot Initial State, Language, Lighting,
Background, Sensor Noise, and Object Layout.
Each shift-specific score pools rollouts across the four suites, and Total is the success rate pooled over all evaluated rollouts across suites and shifts, rather than an unweighted average of the seven shift-specific scores.

For the controlled view-intervention experiments, we independently perturb the
third-person and wrist views while keeping the other visual source unchanged.
Black-out replaces the selected view entirely, whereas graded blur progressively
degrades its visual information.
All evaluation settings, random seeds, and rollout counts are fixed across the
compared methods.


\section{Computational Cost}
\label{app:compute}

Experiments were conducted on NVIDIA A100, RTX~6000 Ada, and H100 GPUs.
We report approximate compute in A100-hours (A100-h), using throughput-based
conversion factors of $1.0\times$, $0.9\times$, and $2.0\times$, respectively.
Costs are estimated from scheduler logs and exclude failed runs and preliminary tests.

\begin{table}[H]
\centering
\caption{\textbf{Approximate computational cost for the primary Qwen3.5-0.8B-OFT experiments.}
Training costs are reported per run and evaluation costs per checkpoint, in A100-hours.}
\label{tab:compute-primary}
\small
\begin{tabular}{lr}
\toprule
\textbf{Run / Evaluation} & \textbf{A100-h} \\
\midrule
\multicolumn{2}{l}{\textit{Training}} \\
\hdashline
SFT (30k steps) & 88 \\
Direct AT (30k steps) & 181 \\
\hdashline
SFT initialization (3,750 steps) & 11 \\
SFT$\rightarrow$AT (+10k) & 49 \\
SFT$\rightarrow$Swap (+10k) & 28 \\
SFT$\rightarrow$AT+Swap (+10k) & 47 \\
\midrule
\multicolumn{2}{l}{\textit{Evaluation}} \\
\hdashline
Original LIBERO (4 suites) & $\sim$3.6 \\
LIBERO-Plus (4 suites, 7 shifts) & $\sim$45 \\
\bottomrule
\end{tabular}

{\footnotesize $^\ast$Estimated from the median of evaluations with complete timing records.}
\end{table}

The SFT-initialized variants share the same 3,750-step SFT checkpoint; its
cost is counted only once in the aggregate compute below.

\begin{table}[H]
\centering
\caption{\textbf{Approximate aggregate computational cost of the experiments.}}
\label{tab:compute-total}
\small
\begin{tabular}{lr}
\toprule
\textbf{Category} & \textbf{A100-h} \\
\midrule
Training & 2,167 \\
Original LIBERO evaluation & 115 \\
LIBERO-Plus evaluation & 1,536 \\
Additional evaluations & 1,669 \\
\midrule
\textbf{Total} & \textbf{5,487} \\
\bottomrule
\end{tabular}
\end{table}

In total, the experiments summarized above required approximately
5,487 A100-h (229 A100-days), with training accounting for approximately
39\% of the total compute.

\section{Probing the Mechanism of View Collapse}
\label{app:view-collapse-mechanism}

Joint Direct AT eventually produces a policy that relies almost exclusively on the \wrist{} view.
We investigate how this direction is selected and how the imbalance develops during training.
Specifically, we test whether the collapse reflects an intrinsic preference for the \wrist{} view or depends on which view is attacked, and examine whether joint AT begins from an asymmetric state.

\noindent\textbf{Attack-target intervention.}
We first intervene on the attacked view during from-scratch AT.
We train two variants that differ only in attack placement:
\texttt{3rd-only-AT} perturbs only the \agent{} image, whereas
\texttt{wrist-only-AT} perturbs only the \wrist{} image.
Both use single-step $\ell_\infty$ PGD with $\epsilon=2/255$ and the 10k-step schedule.
At steps 1000, 2000, and 3000, we evaluate each checkpoint on LIBERO-Object under clean input, \wrist{} black-out, and \agent{} black-out, using 50 trials per task across 10 tasks.

\begin{table}[H]
\centering
\caption{\textbf{Attack placement reverses view reliance.}
Success rates (\%) on LIBERO-Object.
Each policy remains effective when the attacked view is removed, but fails when the unattacked view is removed.}
\label{tab:attack-target-intervention}
\small
\begin{tabular}{lrrrr}
\toprule
Training & Step & Clean & Wrist black-out & 3rd black-out \\
\midrule
\texttt{3rd-only-AT}
    & 1000 & 83.0 & 0.0  & 78.4 \\
    & 2000 & 87.4 & 0.0  & 87.6 \\
    & 3000 & 93.4 & 0.0  & 88.0 \\
\midrule
\texttt{wrist-only-AT}
    & 1000 & 21.4 & 23.2 & 0.0 \\
    & 2000 & 52.2 & 51.6 & 0.0 \\
    & 3000 & 49.2 & 50.0 & 0.0 \\
\bottomrule
\end{tabular}
\end{table}

Table~\ref{tab:attack-target-intervention} shows a mirrored reliance pattern.
When only the \agent{} view is attacked, removing that view has little effect, whereas removing the unattacked \wrist{} view reduces success to zero.
Attacking only the \wrist{} view produces the opposite pattern.
The direction remains stable across all three checkpoints.
Thus, attack placement causally affects which view is suppressed during from-scratch AT, ruling out an unconditional tendency to select the \wrist{} view.
The \wrist{}-reliant variant nevertheless learns faster and reaches higher clean success, suggesting that \wrist{} observations may still offer an optimization or task-information advantage; this advantage alone, however, does not determine the direction of collapse.

\noindent\textbf{Per-view adversarial sensitivity.}
The intervention above explains how attack placement can control the selected view, but joint Direct AT attacks both views.
To examine why it initially moves toward \wrist{} reliance, we measure the loss increase induced by attacking each view separately:
\begin{equation}
    \Delta\ell_v
    =
    \ell_\theta(x+\hat{\delta}_v)-\ell_\theta(x),
    \qquad v\in\{g,w\},
    \label{eq:per-view-adversarial-sensitivity}
\end{equation}
where $g$ and $w$ denote the \agent{} and \wrist{} views.
The perturbation $\hat{\delta}_v$ is generated using the same
implementation as in training, with one step,
$\epsilon=2/255$, step size $\epsilon$, and a random start, but is restricted to view $v$.
Because this is an approximate attack rather than exact inner maximization, individual loss changes can be slightly negative.

We sample mid-episode frames from the LIBERO training data, stratified across the four suites.
Each checkpoint is evaluated on three fixed probe-set draws of 128 examples, giving 384 pooled evaluations.
We report the mean difference
$d=\Delta\ell_g-\Delta\ell_w$, its paired $t$-statistic, and the fraction of examples satisfying $\Delta\ell_g>\Delta\ell_w$.
The initialization checkpoint consists of the pretrained VLM with newly initialized LoRA parameters and action head, before any LIBERO-specific updates.

\begin{table}[H]
\centering
\caption{\textbf{Per-view adversarial sensitivity before and during joint Direct AT.}
Positive $\bar d$ indicates a larger loss increase from perturbing the \agent{} view.
The initial ordering disappears by step 1000 and clearly reverses by steps 2000--3000.}
\label{tab:per-view-adversarial-sensitivity}
\small
\resizebox{0.8\columnwidth}{!}{%
\begin{tabular}{lrrrrr}
\toprule
Checkpoint
& $\overline{\Delta\ell_g}$
& $\overline{\Delta\ell_w}$
& $\bar d$
& $t$
& $\Delta\ell_g>\Delta\ell_w$ \\
\midrule
Initialization
& $0.01610$ & $0.00750$
& $\mathbf{+0.00860}$ & $\mathbf{+21.15}$ & $86\%$ \\
Direct AT, step 1000
& $-0.00002$ & $0.00003$
& $-0.00005$ & $-2.11$ & $45\%$ \\
Direct AT, step 2000
& $0.00022$ & $0.00267$
& $\mathbf{-0.00245}$ & $\mathbf{-9.87}$ & $15\%$ \\
Direct AT, step 3000
& $0.00037$ & $0.00334$
& $\mathbf{-0.00298}$ & $\mathbf{-10.76}$ & $11\%$ \\
\bottomrule
\end{tabular}%
}
\end{table}

\noindent\textbf{An initial asymmetry aligns with the collapse direction.}
Before LIBERO-specific training, perturbing the \agent{} view induces a substantially larger loss increase than perturbing the \wrist{} view:
$\bar d=0.0086$, with the same ordering on 86\% of probe examples.
This asymmetry is already present when joint Direct AT begins and aligns with the subsequent suppression of the \agent{} stream.
Because the measurement includes the newly initialized action head and LoRA parameters, we do not attribute it solely to the pretrained VLM backbone.

The ordering changes rapidly during training.
By step 1000, the mean difference is close to zero.
By steps 2000--3000, perturbing the \wrist{} view induces a clearly larger loss increase, and the magnitude of this reversed imbalance grows.
A companion gradient analysis shows the same transition: the \agent{} share of the total visual input-gradient magnitude decreases from 0.435 at step 1000 to 0.146 at step 2000.
Task success remains approximately zero at both checkpoints and begins to recover only around step 5000.
The shift in view influence therefore emerges before successful task behavior and becomes more pronounced as training proceeds.

\noindent\textbf{Selection followed by amplification.}
The combined results are consistent with a two-stage account of view collapse.
First, the attack-target intervention shows that the attacked view can determine the initial direction of view selection.
Under joint attacks, the initialization-level sensitivity imbalance provides a plausible bias toward suppressing the \agent{} view.
Second, once a \wrist{}-side bias emerges, subsequent training increasingly concentrates gradients and adversarial loss sensitivity on the \wrist{} stream, consistent with amplification of the selected direction.

This interpretation does not treat a larger $\Delta\ell_v$ as evidence that view $v$ is intrinsically harder to robustify.
The quantity reflects the current policy's sensitivity and therefore depends on both the robustness of the extracted features and how strongly the policy uses that view.
As \wrist{} reliance develops, attacking the \wrist{} stream has a larger effect, whereas attacking the increasingly suppressed \agent{} stream has little effect.
Accordingly, the intervention establishes attack placement as a causal factor in view selection, while the initial sensitivity asymmetry and its evolution provide evidence consistent with, but do not fully identify, the subsequent amplification mechanism.

\section{Additional Analyses of Direct Adversarial Training}
\label{app:at-sweep}

\begin{table}[H]
\centering
\caption{\textbf{The shift-specific robustness pattern persists across adversarial threat models and perturbation budgets.}
Across multiple configurations, Direct AT produces pronounced gains on Camera Viewpoint and Sensor Noise, together with mixed gains and degradations across the remaining shifts.
Values are four-suite success rates (\%).}
\label{tab:at-threat-ablation}
\small
\resizebox{\textwidth}{!}{%
\begin{tabular}{lccccccccc}
\toprule
& \multicolumn{1}{c}{\textbf{Original}} & \multicolumn{8}{c}{\textbf{LIBERO-Plus}} \\
\cmidrule(lr){2-2}
\cmidrule(lr){3-10}
\textbf{Training} & \textbf{Total} & \textbf{Camera} & \textbf{Robot} & \textbf{Language} & \textbf{Light} & \textbf{Background} & \textbf{Noise} & \textbf{Layout} & \textbf{Total} \\
\midrule
SFT & 92.7 & 39.8 & 55.7 & 64.4 & 84.8 & 82.2 & 53.3 & 71.7 & 62.7 \\
\midrule

Direct AT ($\ell_\infty$, $\epsilon{=}2/255$)
& 92.1 & 75.7 & 35.5 & 59.3 & 13.0 & 29.8 & 83.0 & 60.7 & 53.8 \\
& \cellcolor{red!5}$\downarrow 0.6$
& \cellcolor{green!32}$\uparrow 35.9$
& \cellcolor{red!22}$\downarrow 20.2$
& \cellcolor{red!8}$\downarrow 5.1$
& \cellcolor{red!50}$\downarrow 71.8$
& \cellcolor{red!42}$\downarrow 52.4$
& \cellcolor{green!28}$\uparrow 29.7$
& \cellcolor{red!14}$\downarrow 11.0$
& \cellcolor{red!12}$\downarrow 8.9$ \\[0.2mm]

Direct AT ($\ell_\infty$, $\epsilon{=}1/255$)
& 92.9 & 88.4 & 36.9 & 69.2 & 68.2 & 66.7 & 86.8 & 68.6 & 69.6 \\
& \cellcolor{green!5}$\uparrow 0.2$
& \cellcolor{green!40}$\uparrow 48.6$
& \cellcolor{red!20}$\downarrow 18.8$
& \cellcolor{green!7}$\uparrow 4.8$
& \cellcolor{red!16}$\downarrow 16.6$
& \cellcolor{red!15}$\downarrow 15.5$
& \cellcolor{green!31}$\uparrow 33.5$
& \cellcolor{red!5}$\downarrow 3.1$
& \cellcolor{green!9}$\uparrow 6.9$ \\[0.2mm]

Direct AT ($\ell_2$, $\epsilon{=}2$)
& 89.9 & 90.0 & 33.9 & 61.6 & 57.5 & 58.2 & 91.3 & 60.3 & 65.6 \\
& \cellcolor{red!6}$\downarrow 2.8$
& \cellcolor{green!42}$\uparrow 50.2$
& \cellcolor{red!24}$\downarrow 21.8$
& \cellcolor{red!6}$\downarrow 2.8$
& \cellcolor{red!25}$\downarrow 27.3$
& \cellcolor{red!23}$\downarrow 24.0$
& \cellcolor{green!35}$\uparrow 38.0$
& \cellcolor{red!13}$\downarrow 11.4$
& \cellcolor{green!5}$\uparrow 2.9$ \\[0.2mm]

Direct AT ($\ell_2$, $\epsilon{=}1$)
& 96.9 & 31.0 & 39.8 & 64.7 & 74.3 & 77.1 & 67.0 & 75.9 & 60.0 \\
& \cellcolor{green!7}$\uparrow 4.2$
& \cellcolor{red!11}$\downarrow 8.8$
& \cellcolor{red!18}$\downarrow 15.9$
& \cellcolor{green!1}$\uparrow 0.3$
& \cellcolor{red!12}$\downarrow 10.5$
& \cellcolor{red!7}$\downarrow 5.1$
& \cellcolor{green!15}$\uparrow 13.7$
& \cellcolor{green!7}$\uparrow 4.2$
& \cellcolor{red!6}$\downarrow 2.7$ \\

\bottomrule
\end{tabular}%
}
\end{table}

\paragraph{Threat-model and budget ablations.}
To examine whether the highly shift-dependent behavior of Direct AT is specific to the main $\ell_\infty$, $\epsilon{=}1/255$ configuration, we additionally evaluate multiple perturbation norms and budgets.
Table~\ref{tab:at-threat-ablation} reports the complete seven-axis LIBERO-Plus results.

The results show that this shift-specific behavior persists across multiple adversarial settings, but is not universal.
Increasing the $\ell_\infty$ budget to $\epsilon{=}2/255$ produces a more severe trade-off: Camera Viewpoint and Sensor Noise remain substantially improved, while several other shifts degrade much more strongly. A similar pattern also appears under $\ell_2$ AT with $\epsilon{=}2$, which again yields large gains on Camera Viewpoint and Sensor Noise at the cost of degradations on other shifts.
In contrast, the weaker $\ell_2$, $\epsilon{=}1$ configuration does not exhibit the same Camera Viewpoint gain and shows a qualitatively different generalization profile.
Thus, the shift-specific effects of Direct AT are reproducible across
different threat models and perturbation budgets, while their magnitude and failure profile depend strongly on the AT configuration.

\begin{figure}[t]
    \centering
    \begin{subfigure}[t]{1.0\linewidth}
        \centering
        \includegraphics[width=\textwidth]{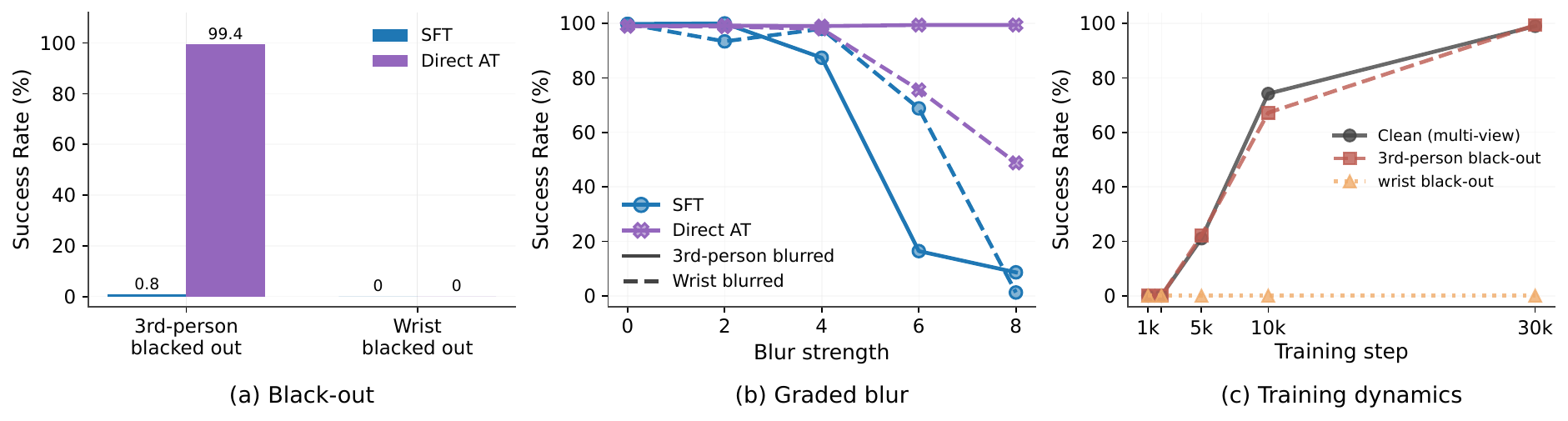}
        \caption{Direct AT ($\ell_\infty$, $\epsilon{=}2/255$)}
    \end{subfigure}
    \hfill
    \begin{subfigure}[t]{0.6\linewidth}
        \centering
        \includegraphics[width=\textwidth]{
        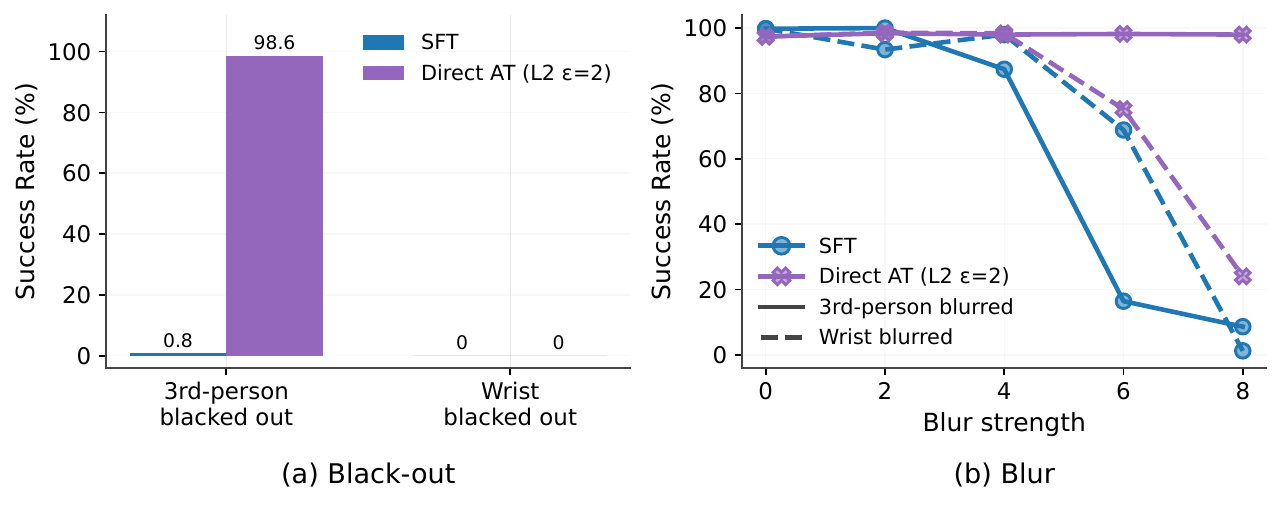}
        \caption{Direct AT ($\ell_2$, $\epsilon{=}2$)}
    \end{subfigure}
    \hfill
    \begin{subfigure}[t]{0.6\linewidth}
        \centering
        \includegraphics[width=\textwidth]{
        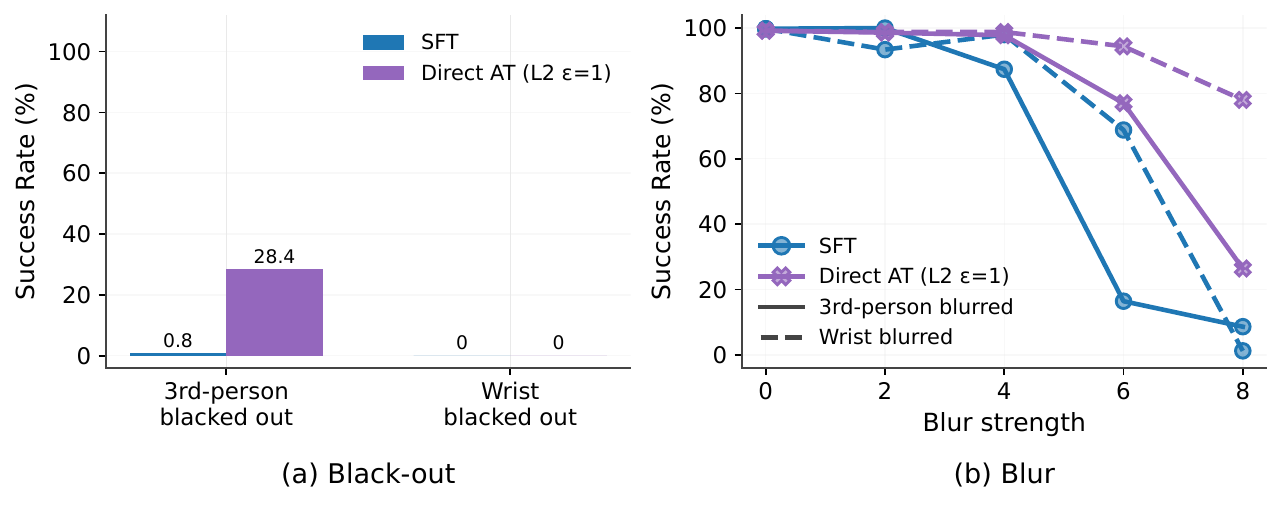}
        \caption{Direct AT ($\ell_2$, $\epsilon{=}1$)}
    \end{subfigure}
    \caption{\textbf{View-intervention results for additional Direct AT configurations.}
    The two settings that substantially improve Camera Viewpoint exhibit
    near-complete wrist-dominant collapse, whereas $\ell_2$, $\epsilon{=}1$
    shows a qualitatively different pattern with greater sensitivity to
    third-person degradation.}
    \label{fig:at-config-view-reliance}
\end{figure}

\paragraph{View reliance across threat models.}
\label{app:at-view-collapse}
We next test whether the view collapse identified for the main $\ell_\infty$, $\epsilon{=}1/255$ configuration also appears under alternative AT settings.
Figure~\ref{fig:at-config-view-reliance} shows the same near-complete wrist-dominant collapse for $\ell_\infty$, $\epsilon{=}2/255$ and $\ell_2$, $\epsilon{=}2$, both of which substantially improve Camera Viewpoint.
In contrast, $\ell_2$, $\epsilon{=}1$, which does not improve Camera Viewpoint, shows a qualitatively different reliance pattern and is more sensitive to third-person degradation under graded blur.
Together, these results show that the severity of single-view reliance depends on the AT configuration, with several settings reaching near-complete collapse.

\begin{figure}[t]
    \centering
    \includegraphics[width=\linewidth]{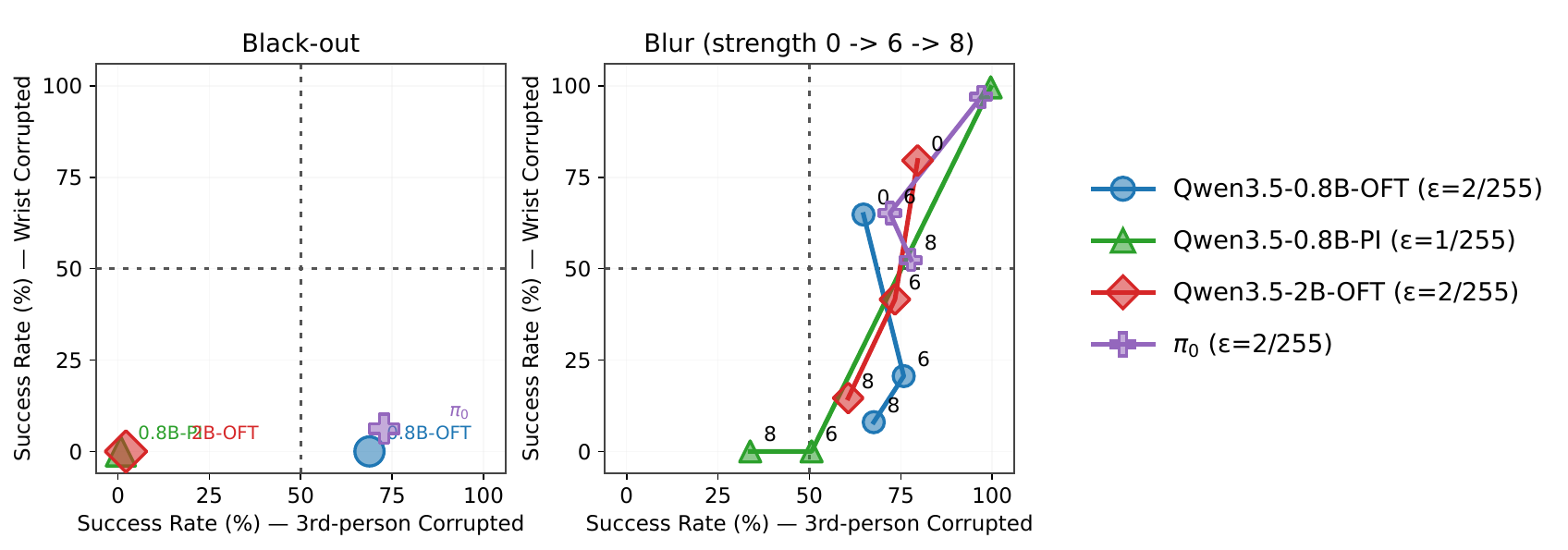}
    \caption{\textbf{View reliance after 3k steps of Direct AT across model configurations.}
    Success rates under third-person-view corruption (horizontal axis) and wrist-view corruption (vertical axis).
    Under graded blur, all configurations show greater sensitivity to wrist-view degradation, while black-out responses vary.
    Perturbation budgets are shown in the legend.}
    \label{fig:early-at-view-reliance}
\end{figure}

\paragraph{View reliance across model configurations.}
\label{app:cross-model-view-reliance}
We extend the view-intervention analysis to additional model configurations after a short 3k-step Direct AT.
Under graded blur, all tested configurations are more sensitive to wrist-view degradation than to third-person-view degradation, indicating wrist-biased reliance (Figure~\ref{fig:early-at-view-reliance}).
Black-out responses vary across models, and not all configurations exhibit the near-complete view collapse observed in our primary setting.
These results suggest that wrist-biased reliance under Direct AT extends beyond the primary model, although its severity varies across configurations.

\section{View Swap versus Matched View Dropout}
\label{sec:swap-dropout-ablation}

We further test whether simply removing one visual source is sufficient to obtain the benefit of View Swap. As a matched baseline, we use View Dropout, which blacks out one randomly selected view at the same 25\% training frequency. Both variants use the same SFT initialization and adversarial-training configuration; only the view intervention is changed.

Our hypothesis is that View Swap provides a more informative reliability intervention than View Dropout. A blacked-out view is an obvious out-of-distribution signal that the source is unavailable, allowing the policy to detect and ignore it directly. In contrast, a swapped view remains visually plausible and preserves the appearance statistics of the corresponding camera, but its content is inconsistent with the current task context. The policy must therefore learn that an apparently valid view can still be unreliable.

\begin{table}[H]
\centering
\caption{\textbf{View Swap versus matched View Dropout on \liberop{}.} Both variants use the same SFT initialization, adversarial-training configuration, and view-intervention frequency. Values are four-suite success rates (\%).}
\label{tab:swap-dropout}
\small
\setlength{\tabcolsep}{4pt}
\resizebox{\textwidth}{!}{%
\begin{tabular}{lccccccccc}
\toprule
& \multicolumn{1}{c}{\textbf{Original}} & \multicolumn{8}{c}{\textbf{LIBERO-Plus}} \\
\cmidrule(lr){2-2}\cmidrule(lr){3-10}
\textbf{Method} & \textbf{Total} & \textbf{Camera} & \textbf{Robot} & \textbf{Language} & \textbf{Light} & \textbf{Background} & \textbf{Noise} & \textbf{Layout} & \textbf{Total} \\
\midrule
SFT$\rightarrow$AT+Dropout & \textbf{97.3} & 63.6 & 60.4 & \textbf{72.3} & 86.1 & \textbf{91.0} & 85.1 & 76.9 & 75.4 \\
SFT$\rightarrow$AT+Swap & 96.5 & \textbf{84.1} & \textbf{65.2} & 68.4 & \textbf{88.8} & 87.7 & \textbf{88.3} & \textbf{79.5} & \textbf{79.7} \\
\midrule
$\Delta$ (Swap $-$ Dropout) & -0.8 & +20.5 & +4.8 & -3.8 & +2.7 & -3.3 & +3.1 & +2.6 & +4.3 \\
\bottomrule
\end{tabular}%
}
\end{table}

As shown in Table~\ref{tab:swap-dropout}, View Swap provides stronger overall robustness than matched View Dropout while maintaining comparable original-task performance. This result supports our hypothesis that training on plausible-but-unreliable views provides a more useful signal than exposing the policy only to completely missing visual inputs.

\section{Comparison with Prior Methods on LIBERO-Plus}
\label{app:sota-libero-plus}

To contextualize our results, we compare our primary Qwen3.5-0.8B-OFT model with representative methods evaluated on LIBERO-Plus. We distinguish direct VLM-to-VLA adaptation from models with prior robot-action pretraining; all models except OpenVLA-OFT + PT are evaluated zero-shot on LIBERO-Plus without training on its distribution shifts.

\begin{table}[H]
\centering
\caption{\textbf{Comparison with prior methods on LIBERO-Plus.} Values are four-suite pooled success rates (\%). Bold indicates the best result among models directly adapted from a VLM using only standard LIBERO training data.}
\label{tab:sota-libero-plus}
\small
\resizebox{\textwidth}{!}{%
\begin{tabular}{l:c:cccccccc}
\toprule
\textbf{Method} & \textbf{Size} & \textbf{Camera} & \textbf{Robot} & \textbf{Language} & \textbf{Light} & \textbf{Background} & \textbf{Noise} & \textbf{Layout} & \textbf{Total} \\
\midrule
\multicolumn{10}{l}{\textit{Direct VLM-to-VLA adaptation on standard LIBERO}} \\
Qwen3-VL-OFT & 4B & 47.0 & 60.1 & \textbf{87.0} & \textbf{96.3} & \textbf{95.3} & 73.1 & 79.2 & 75.0 \\
Qwen3-VL-PI & 4B & 64.3 & 57.2 & 82.8 & 94.2 & 94.0 & 79.6 & 78.2 & 77.0 \\
Qwen3.5-0.8B-OFT (SFT) & 0.8B & 38.8 & 55.7 & 71.0 & 88.9 & 88.5 & 59.1 & 69.8 & 65.3 \\
\textbf{Qwen3.5-0.8B-OFT (SFT$\rightarrow$AT+Swap)} & 0.8B & \textbf{84.1} & \textbf{65.2} & 68.4 & 88.8 & 87.7 & \textbf{88.3} & \textbf{79.5} & \textbf{79.7} \\
\midrule
\multicolumn{10}{l}{\textit{Robot-pretrained VLAs finetuned on standard LIBERO}} \\
OpenVLA-OFT~\citep{fei2026liberoplus} & 7B & 56.4 & 31.9 & 79.5 & 88.7 & 93.3 & 75.8 & 74.2 & 69.6 \\
$\pi_0$ (public repro.)$^\ddagger$ & 3.3B & 17.3 & 6.7 & 63.6 & 86.5 & 82.1 & 17.6 & 74.2 & 46.3 \\
$\pi_0$-Fast~\citep{fei2026liberoplus} & 3.3B & 65.1 & 21.6 & 61.0 & 73.2 & 73.2 & 74.4 & 68.8 & 61.6 \\
ABot-M0~\citep{yang2026abot} & 4B & 60.4 & 67.9 & 86.4 & 96.2 & 91.6 & 86.4 & 82.6 & 80.5 \\
$\pi_{0.5}$~\citep{black2025pi05} & 3.3B & 78.4 & 73.6 & 80.8 & 96.2 & 94.1 & 89.0 & 84.5 & 84.4 \\
Qwen-RobotManip~\citep{yuan2026qwen} & 4B & 87.2 & 75.5 & 85.6 & 96.6 & 97.7 & 97.7 & 87.3 & 89.0 \\
\midrule
\multicolumn{10}{l}{\textit{Training with generalized LIBERO data}} \\
OpenVLA-OFT + PT$^\dagger$~\citep{fei2026liberoplus} & 7B & 92.8 & 30.3 & 85.8 & 94.9 & 93.9 & 89.3 & 77.6 & 79.5 \\
\bottomrule
\end{tabular}%
}
\vspace{1pt}
{\footnotesize\raggedright Model size denotes the nominal model/backbone scale; auxiliary action-head parameters are not consistently reported. $^\dagger$PT denotes post-training with more than 20K additional generalized trajectories generated through the LIBERO-Plus pipeline. LIBERO-Plus results for $\pi_{0.5}$ are taken from \citet{yuan2026qwen}. $^\ddagger$For $\pi_0$, we use the public full-benchmark reproduction with the official OpenPI checkpoint~\citep{liberoplus-pi0-repro}, which reports 17.6\% Sensor Noise versus 79.0\% in the original leaderboard. \par}
\end{table}

Qwen3-VL-OFT and Qwen3-VL-PI results are taken from the StarVLA LIBERO-Plus evaluation. Our 0.8B direct VLM-to-VLA baseline improves from 65.3\% to 79.7\% overall success with SFT$\rightarrow$AT+Swap. The resulting model achieves the highest average among the direct VLM-to-VLA models in this comparison, with particularly strong Camera Viewpoint (84.1\%) and Sensor Noise (88.3\%) robustness, while remaining competitive with several substantially larger robot-pretrained VLAs.

\section{Visualization of Training Perturbations}
\label{app:perturbation-visualization}

\begin{figure}[t]
\centering
\begin{subfigure}[t]{0.85\textwidth}
    \centering
    \includegraphics[width=\linewidth]{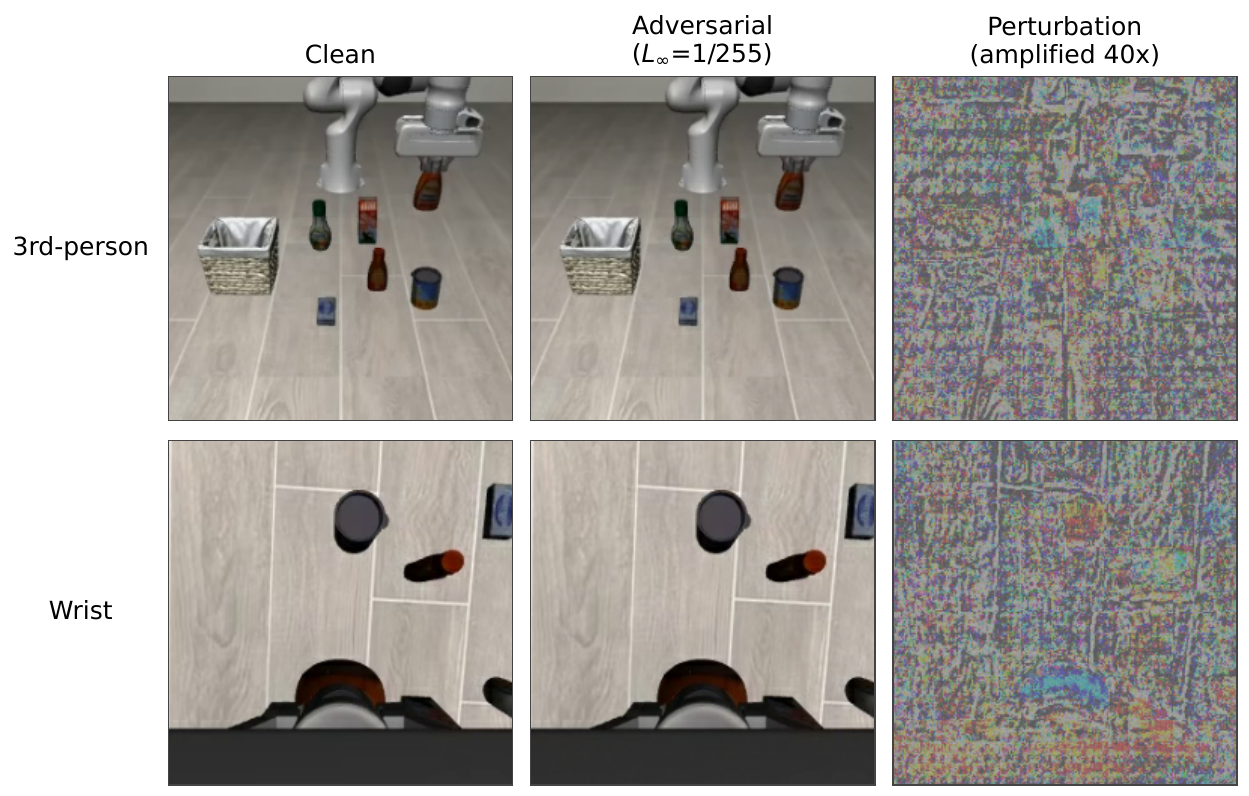}
    \caption{\textbf{Adversarial perturbation.} Clean and adversarial observations under 1-step PGD with $\ell_\infty$, $\epsilon{=}1/255$. The right column visualizes the perturbation amplified by $40\times$.}
    \label{fig:adv-example}
\end{subfigure}
\hfill
\vspace{10pt}
\begin{subfigure}[t]{0.85\textwidth}
    \centering
    \includegraphics[width=\linewidth]{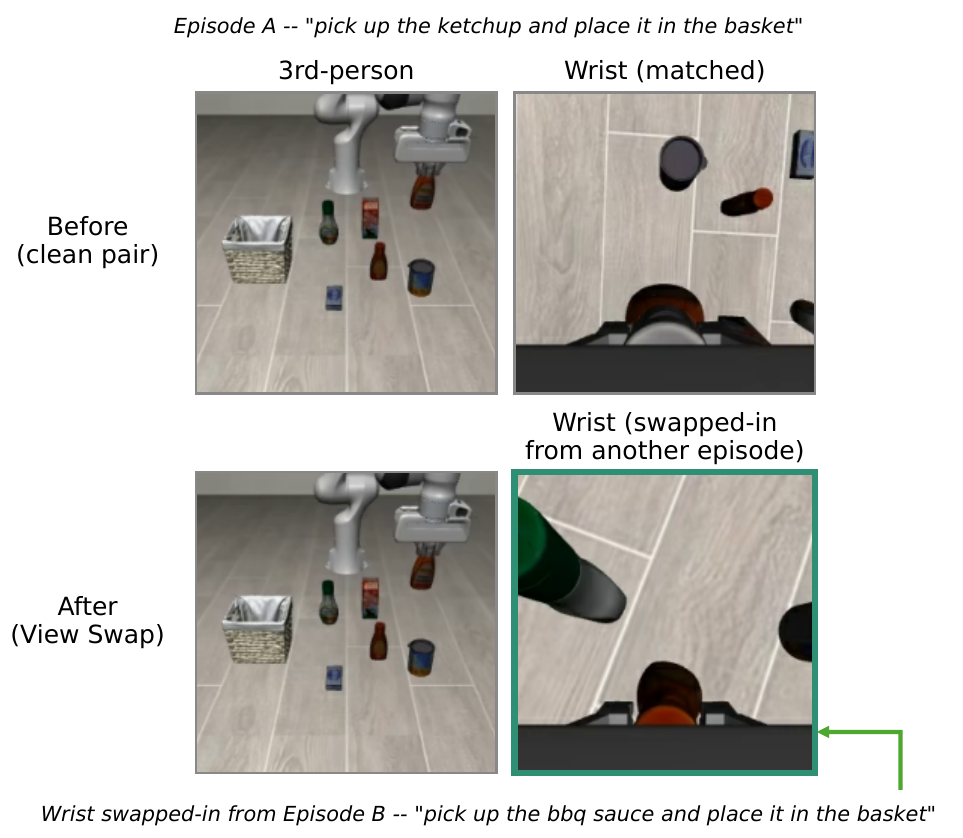}
    \caption{\textbf{View Swap.} The third-person observation is kept unchanged, while the wrist observation is replaced by a frame from another sample, producing a mismatched multi-view pair.}
    \label{fig:swap-example}
\end{subfigure}
\caption{\textbf{Visualization of the training perturbations.} Adversarial training perturbs pixels within each view, while View Swap perturbs the correspondence across views.}
\label{fig:training-perturbations}
\end{figure}

Figure~\ref{fig:training-perturbations} provides qualitative examples of the two input perturbations used in our training. Pixel-space AT applies small adversarial perturbations to both camera views, whereas View Swap explicitly breaks the correspondence between views by replacing one view with a frame from another sample.

\end{document}